# Performance Evaluation of SIFT Descriptor against Common Image Deformations on Iban Plaited Mat Motifs

Silvia Joseph[1], Irwandi Hipiny[2] and Hamimah Ujir[3]

*[1]Universiti Malaysia Sarawak,* silvia.joe8@gmail.com
*[2]Universiti Malaysia Sarawak,* mhihipni@unimas.my
*[3]Universiti Malaysia Sarawak,* uhamimah@unimas.my

Borneo indigenous communities are blessed with rich craft heritage. One such examples is the Iban's plaited mat craft. There have been many efforts by UNESCO and the Sarawak Government to preserve and promote the craft. One such method is by developing a mobile app capable of recognising the different mat motifs. As a first step towards this aim, we presents a novel image dataset consisting of seven mat motif classes. Each class possesses a unique variation of chevrons, diagonal shapes, symmetrical, repetitive, geometric and non-geometric patterns. In this study, the performance of the Scale-invariant feature transform (SIFT) descriptor is evaluated against five common image deformations, i.e., zoom and rotation, viewpoint, image blur, JPEG compression and illumination. Using our dataset, SIFT performed favourably with test sequences belonging to Illumination changes, Viewpoint changes, JPEG compression and Zoom + Rotation. However, it did not performed well with Image blur test sequences with an average of 1.61% retained pairwise matching after blurring with a Gaussian kernel of 8.0 radius.

## 1.0    Introduction

Sarawak, the Land of Hornbill, is the largest Malaysian state with the most number of indigenous tribes. When it comes to cultural art craft, Sarawak is blessed with diverse backgrounds and traditions which produce a plethora of patterns, motifs and designs for creating unique and beautiful products. Sarawak is famous with its traditional art forms, or rather handicrafts, made by the indigenous people with materials collected from the rich rainforests. A handicraft product is defined by the Malaysian Handicraft Development Corporation as any products that has artistic or traditional cultural attractions and is the result of a process that depends solely or partly to the skill of hand (Tiing, 2013).

In this study, we focus on Iban's mat plaiting motifs plaited by one of the indigenous communities in Sarawak, i.e., Iban Batang Ai from Rumah Bintung, Lubuk Antu. As there is no written documentation, it is rather vague on how the Ibans came to learn mat plaiting skill. The skill is considered as one of the oldest traditions according to Selleto (2012) amongst the Ibans, with the plaiting of plain mats considered to be of the same status as knitting in traditional European societies. Previously, it was deemed as an essential asset to an Iban girl that she would be unfailingly taught the technique upon reaching a teenage age by her mother and grandmother.

Normally, mat plaiting activity is performed in the common gallery or 'ruai' of the longhouse. Plaited mats were often ornamented with patterned, abstract design with distinctive ethnic and regional identifiers. The mats' common denominator is the plaiting technique, plaited without a loom and historically made only of natural fibres such as 'bemban' or known as *Donax arundastrum*, a white flowering reed of the arrow-root family which grows in swampy places (Durin, 2014). An example of colourless 'bemban' mat is shown in Figure 1(a). According to Sentance (2007), the difficulty of sourcing traditional materials has led craftsmen to use synthetic strips instead, see Figure 1(b).

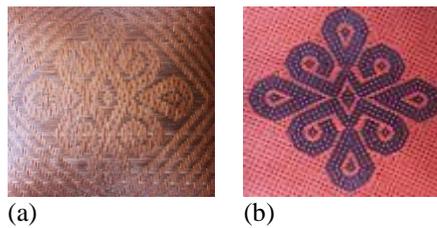

(a)　　　　　(b)

Figure 1: An example of (a) a colourless 'bemban' mat and (b) a synthetic mat. Both contains the same motif.

The standard rectangular plaited mat, see Figure 2, consists of an outer edge and two or more inner frames of transposed twill with decorative small motifs as fillers. The centre of the mat, i.e., "pulau", may contain a single main motif (one instance or repetitive) or more.

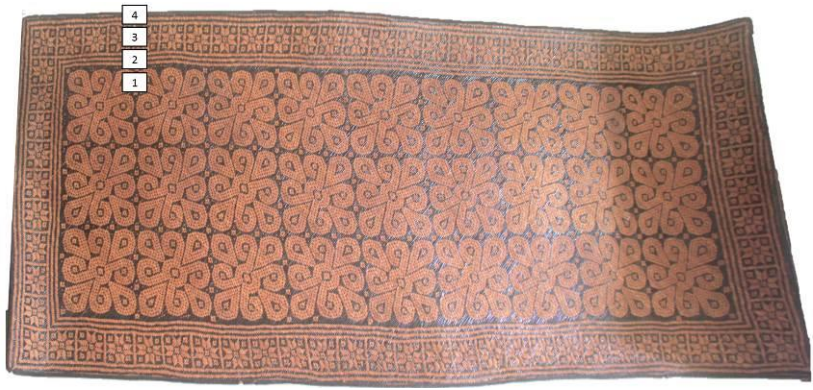

Figure 2. An example of a "Buah Tungku" plaited mat with a single large "pulau" (1); the decorative edge (3) is positioned between two strengthening rows of transposed twill, i.e., "sapa" (2, 4).

On a single mat, each main motif has a unique philosophical meaning whilst smaller motifs, acting as fillers, do not represent any. Iban plaited mat motifs are named after natural phenomenon, mostly stylised beyond recognition (Selleto, 2012). The names fall into loose categories (i.e., classes) such as plants, animal, firmaments, faerie and things. Each mat motif is named as a "buah" (literally a fruit). The motifs are designed through a human cognitive process that represent an artistic interpretation of nature in geometric shapes.

This work evaluates Lowe's (1999) Scale-invariant feature transform (SIFT) descriptor against several image deformations. SIFT is a robust and invariant gradient-based descriptor for describing local feature vectors found inside a 2D image. Iban's plaited mat motifs generally contains rich textures, i.e., strong gradients, which fits the strengths of SIFT.

**2.0   Literature review**

In this section, we review several Computer vision methods related to the task of image matching and retrieval. According to Balntas et al. (2017), local feature descriptors make use of low level features, such as edges and gradients, to increase robustness against image deformations. In (Zhang et al., 2012), they mentioned that due to the unstructured array of pixels inside an image, the first step in semantic understanding is to extract a set of efficient and effective visual features from those pixels. Several feature detectors and descriptors had been proposed in literature with a variety of definitions to what kind of distinctive low-level features inside an image is potentially interesting. An excellent review of existing methods can be found here (Tuytelaars and Mikoljczyk, 2008).

Selected applications of local feature descriptor in Computer-vision based tasks are discussed next. Madzin et al. (2014) focused on extracting useful information from images of multiple modalities for clinical purposes. They evaluated several local descriptors and had observed that by using a local descriptor, there is no need for a segmentation step to exclude background information. According to (Hassaballah et al., 2016), for certain applications such as camera calibration, image classification, image retrieval and object tracking, it is imperative for the feature descriptor to be robust against changes to brightness and viewpoint as well as image distortions. In (Hipiny, 2013), SIFT is used to describe egocentric image regions fixated by a subject during the performance of a task. The resulting bag-of-features are then used to match the image regions. A similar bag-of-features approach was implemented in (Hipiny et al., 2019) to match sea turtle carapaces using SIFT features. Willy et al. (2013) evaluated the performance of two popular invariant feature descriptors, i.e., Lowe's (1999) Scale-Invariant Feature Transform (SIFT) and Bay's (2008) Speeded-Up Feature Transform (SURF) on "Songket Palembang", a traditional woven cloth from Indonesia. They used both descriptors to classify each motif and had found that SIFT performs better in images with noise density less than 11%.

**3.0   Methodology**

There are three main stages involved in our evaluation of SIFT descriptor against common image deformations. The stages are image acquisition, pre-processing and performance evaluation.

### 3.1 Image acquisition

The plaited mat motif images were captured using a single camera in different illumination settings, rotation + zoom values and viewing angles.

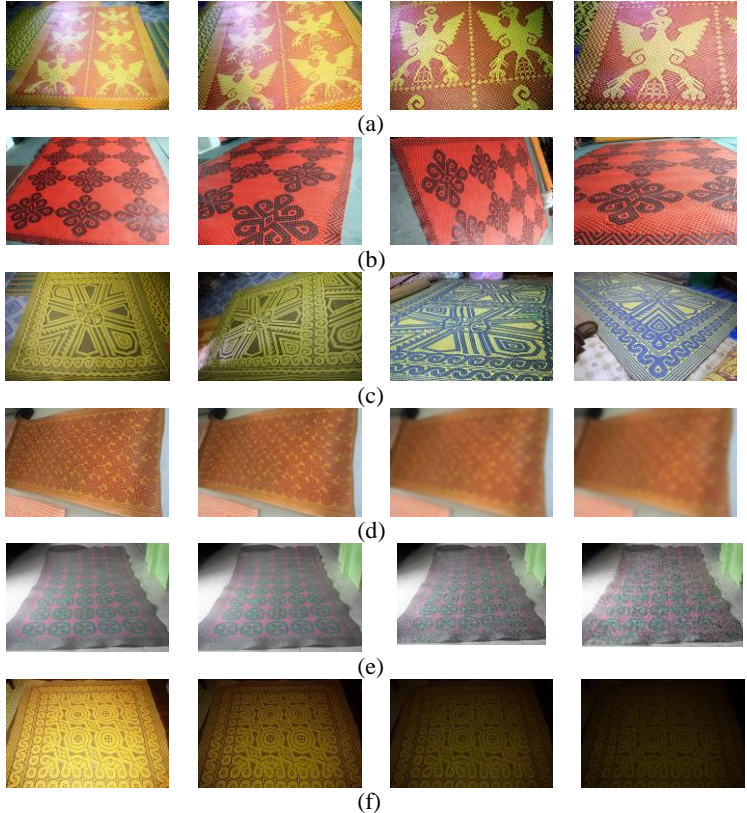

Figure 3. Sample images from our dataset used for (a) **Zoom + Rotation**: "Buah Burung Kenyalang"; **Viewpoint changes**: (b) "Buah Tungku" and (c) "Buah Phancar Matahari"; (d) **Image blur**: "Buah Tampang Keladi Anyut"; (e) **JPEG compression**: "Buah Engkerumung"; and (f) **Illumination changes**: "Buah Bulan Sebayan".

In total, we had collected twenty motif classes. For the purpose of this initial work, we only include seven selected motifs, see Figure 3. The dataset (CC-BY-4.0) is available at the following link: www.fcsit.unimas.my/research/iban-mat-motifs.

### 3.2 Pre-processing

Since we plan to evaluate the SIFT descriptor against illumination and viewpoint changes as well as different zoom + rotation values, we collected images of each mat to reflect each deformation at varying intensities. We then digitally transformed a selected image of each mat to simulate the JPEG compression and blurring effect at varying intensities. The images' original resolution of 5184 x 3456 pixels was also reduced to 800 x 355. Figure 3 shows sample images from our dataset used to evaluate (a) Zoom + Rotation; (b) and (c) Viewpoint changes, (d) Image blur, (e) JPEG compression, and (f) illumination changes. Each test sequence contains 5 images of a gradual geometric and photometric transformation. In total, our dataset contains 7 classes x 5 image deformations x 5 levels = 175 images.

For the Zoom + Rotation test sequences, images were obtained by rotating the camera around its optical axis in the range of 30° to 45° whilst varying the camera's zoom levels. For the Viewpoint test sequences, the camera position were varied from a fronto-parallel view to one with a significant fore shorting at approximately 50° to 60° from the camera. As for the Image blur test sequences, the Gaussian kernel was applied to the original image with a radius ranging from 0.0 to 8.0. For the illumination test sequences, the gradual decrease of lighting were obtained by varying the camera aperture. The JPEG compression test sequences were generated by changing the compression parameter with a value ranging from 100% to 75%.

### 3.3  Performance evaluation

In our experiment, we used each main motif as a template image against query images. The number of positive SIFT matches between the two images is used to determine the quality of the matching. We assume that the higher the number of positive matches correlates to a better matching result. The aim is to discover how well does the SIFT descriptor performed against the five image deformations.

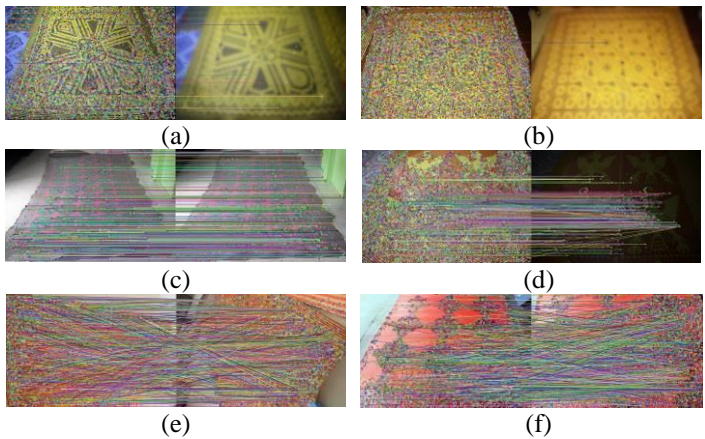

(a)   (b)

(c)   (d)

(e)   (f)

Figure 4. Visualised matched keypoints for (a) **Image blur**: "Buah Phancar Matahari", with threshold set at 0.2; (b) **Image blur**: "Buah Bulan Sebayan", with threshold set at 0.2; (c) **JPEG compression**: "Buah Engkerumung", with threshold set at 0.4; (d) **Illumination changes**: "Buah Burung Kenyalang", with threshold set at 0.6; (e) **Viewpoint changes**: "Buah Tampang Keladi Anyut", with threshold set at 0.8; and (f) **Zoom + Rotation**: "Buah Tungku", with threshold set at 0.8.

We set a range of threshold criterion values for the SIFT feature matching at 0.20, 0.40, 0.60 and 0.80. An identical pair would return a distance score of 0.0.

## 4.0 Results and Discussion

In general, SIFT is quite capable of finding pairwise matches for all five image deformations, even after intense transformation. Nevertheless, as seen in Figure 4, the returned positive matches may consist of false positives. This problem can be solved by estimating the fundamental matrix between the pair of images using RANSAC to reject spurious matches.

Table 1. Selected results from our experiment. The first column represents the image deformation and the intensity level (with a value of 1 to 5, with '1' indicating zero deformation and '5' indicating the most intense deformation). The second column represents the percentage of retained positive matches as compared with the number of detected SIFT features inside the source image.

| Symetrical motif (Buah Phancar Matahari) | % of retained positive matches | Geometric motif (Buah Bulan Sebayan) | % of retained positive matches |
|---|---|---|---|
| Blur-2_sift 0.2 | 3.08% | Blur-2_sift 0.2 | 2.48% |
| Blur-5_sift 0.2 | 0.05% | Blur-5_sift 0.2 | 0.02% |
| Blur-2_sift 0.8 | 15.13% | Blur-2_sift 0.8 | 14.81% |
| Blur-5_sift 0.8 | 0.71% | Blur-5_sift 0.8 | 0.13% |
| Viewpoint-2_sift 0.2 | 0.00% | Viewpoint-2_sift 0.2 | 0.00% |
| Viewpoint-5_sift 0.2 | 0.00% | Viewpoint-5_sift 0.2 | 0.00% |
| Viewpoint-2_sift 0.8 | 5.97% | Viewpoint-2_sift 0.8 | 6.94% |
| Viewpoint-5_sift 0.8 | 8.28% | Viewpoint-5_sift 0.8 | 9.13% |
| Compression-2_sift 0.2 | 57.34% | Compression-2_sift 0.2 | 27.46% |
| Compression-5_sift 0.2 | 57.34% | Compression-5_sift 0.2 | 1.47% |
| Compression-2_sift 0.8 | 81.86% | Compression-2_sift 0.8 | 65.54% |
| Compression-5_sift 0.8 | 81.86% | Compression-5_sift 0.8 | 24.23% |
| Light-2_sift 0.2 | 7.04% | Light-2_sift 0.2 | 15.50% |
| Light-5_sift 0.2 | 2.17% | Light-5_sift 0.2 | 4.16% |
| Light-2_sift 0.8 | 40.66% | Light-2_sift 0.8 | 63.41% |
| Light-5_sift 0.8 | 24.50% | Light-5_sift 0.8 | 27.69% |
| Zoom_Rotation-2_sift 0.2 | 0.00% | Zoom_Rotation-2_sift 0.2 | 0.00% |
| Zoom_Rotation-5_sift 0.2 | 0.01% | Zoom_Rotation-5_sift 0.2 | 0.00% |
| Zoom_Rotation-2_sift 0.8 | 8.96% | Zoom_Rotation-2_sift 0.8 | 7.01% |
| Zoom_Rotation-5_sift 0.8 | 7.51% | Zoom_Rotation-5_sift 0.8 | 5.94% |

| Non-geometric motif (Buah Burung Kenyalang) | % of retained positive matches | Repetitive motifs (Buah Tampang Keladi Anyut) | % of retained positive matches |
|---|---|---|---|
| Blur-2_sift 0.2 | 2.62% | Blur-2_sift 0.2 | 3.51% |
| Blur-5_sift 0.2 | 0.13% | Blur-5_sift 0.2 | 0.03% |
| Blur-2_sift 0.8 | 13.60% | Blur-2_sift 0.8 | 14.02% |
| Blur-5_sift 0.8 | 0.83% | Blur-5_sift 0.8 | 5.01% |
| Viewpoint-2_sift 0.2 | 0.01% | Viewpoint-2_sift 0.2 | 0.01% |
| Viewpoint-5_sift 0.2 | 0.01% | Viewpoint-5_sift 0.2 | 0.00% |
| Viewpoint-2_sift 0.8 | 7.76% | Viewpoint-2_sift 0.8 | 7.02% |
| Viewpoint-5_sift 0.8 | 4.87% | Viewpoint-5_sift 0.8 | 8.66% |
| Compression-2_sift 0.2 | 46.21% | Compression-2_sift 0.2 | 46.21% |
| Compression-5_sift 0.2 | 46.21% | Compression-5_sift 0.2 | 46.21% |
| Compression-2_sift 0.8 | 78.76% | Compression-2_sift 0.8 | 78.28% |
| Compression-5_sift 0.8 | 78.28% | Compression-5_sift 0.8 | 78.28% |
| Light-2_sift 0.2 | 11.03% | Light-2_sift 0.2 | 54.30% |
| Light-5_sift 0.2 | 0.10% | Light-5_sift 0.2 | 1.47% |
| Light-2_sift 0.8 | 38.63% | Light-2_sift 0.8 | 78.26% |
| Light-5_sift 0.8 | 4.77% | Light-5_sift 0.8 | 14.31% |
| Zoom_Rotation-2_sift 0.2 | 0.01% | Zoom_Rotation-2_sift 0.2 | 0.00% |
| Zoom_Rotation-5_sift 0.2 | 0.00% | Zoom_Rotation-5_sift 0.2 | 0.00% |
| Zoom_Rotation-2_sift 0.8 | 10.43% | Zoom_Rotation-2_sift 0.8 | 7.55% |
| Zoom_Rotation-5_sift 0.8 | 6.97% | Zoom_Rotation-5_sift 0.8 | 7.69% |

At the most lax distance threshold criterion of 0.80, we obtained the following results. For Image blur, on average, the retained number of positive matches is 1.61% of the number of detected SIFT features inside the source image (of zero deformation). For Viewpoint changes, on average, the retained positive matches is 7.38% of the number of detected SIFT features in the source image (of zero deformation). For JPEG compression, on average, the retained positive matches is 61.53% of the number of detected SIFT features in the source image (of zero deformation). For Illumination/Light changes, on average, the retained positive matches is 17.87% of the number of detected SIFT features in the source image (of zero deformation). For Zoom + Rotation, on average, the retained positive matches is 8.20% of the number of detected SIFT features in the source image (of zero deformation). As we increase the threshold criterion value, the number of positive matches decreases dramatically.

From the results, SIFT performed very well with the JPEG compression test sequences with very minimal reduction of positive matches even at a high compression rate. SIFT had also performed quite well with the Illumination test sequences. This is as expected as SIFT is based on gradient magnitudes rather than the actual values. As for the Viewpoint and Zoom + rotation test sequences, SIFT is able to return pairwise matching, but the number is greatly reduced from the number of features detected inside the source image (with zero deformation). For our dataset of Iban's plaited mat motifs, SIFT performed poorly on the Image blur test sequences with an average of 1.61% retained pairwise matching.

## 5.0     Conclusion and future

The purpose of this study is to evaluate the fittingness of the SIFT descriptor to describe images of plaited mat motifs for indexing and retrieval. Based on our results, SIFT is shown to be capable of finding pairwise matching, even after a large image deformation. However, the results can be improved further by using RANSAC to exclude spurious matches as well as removing the background clutter (which produces false SIFT features) from the images. We would also like explore the use of colour and edge-based descriptor (Joseph et al., 2017) to describe the Iban mat motifs. Another option is to implement multiple/modular voting scheme (Ujir and Spann 2014; Ujir et al., 2014) to further reduce the false positive matches.

**Acknowledgments**

We would like to thank the craftswomen from Iban long house of Rumah Bintung, Lubuk Antu, Sarawak that willing to share and giving the permission to photographs their craft works of plaited mat.

**References**
Tiing, C.L. (2013). *Sarawak handicrafts: Preserving a heritage threatened by extinction*. Retrieved from http://www.theborneopost.com
Selleto, B. (2012). *Plaited Arts From The Borneo Rainforest.* NUSS Press, Singapore.
Durin, A. (2014). *Tikar Bemban.* Universiti Malaysia Sarawak.
Sentance, B. (2007). *Basketry A World Guide to Traditional Techniques.* Thames & Hudson.

Lowe, D.G. (1999). *Object recognition from local scale-invariant features*. The 7th IEEE International Conference on Computer Vision, Vol 2, pp. 1150–1157.

Balntas, V., Karel, L., Vedaldi, A., and Mikolajczyk, K. (2017). *HPatches: A Benchmark and evaluation of Handcrafted and Learned Local Descriptor*. Computer Vision and Pattern Recognition.

Zhang, D., Islam, M.M., and Lu, G. (2012). *A Review On Automatic Image Annotation Techniques*. Pattern Recognition, pp. 346-362.

Tuytelaars, T. and Mikolajczyk, K. (2008). *Local invariant feature detectors: a survey*. In Journal Foundations and Trends in Computer Graphics and Vision, Vol. 3, pp. 177-280.

Madzin, H., Mohamed, N.S, and Zainuddin R. (2014). *Analysis of Visual Features in Local Descriptor for Multi-Modality Medical Image*. The International Arab Journal of Information Technology, Vol. 11, 468-475.

Hassaballah, M., Abdelmgeid A.A., and Alshazly, H.A. (2016). *Image Features Detection, Description And Matching*. Switzerland: Springer International Publishing Switzerland

Hipiny, I., Ujir, H., Mujahid, A., and Yahya, N.K. (2019). *Towards Automated Biometric Identification of Sea Turtles (Chelonia mydas)*. To appear in International Journal of Pure and Applied Mathematics, Kuwait, ISSN 13118080.

Willy D., Noviyanto, A. and Arymurthy A.A. (2013). *Evaluation of SIFT and SURF Features in the Songket Recognition*. In ICACSIS, Bali.

Bay, H., Tuytelaars, T. and Van Gool, L. (2006). *Surf: Speeded up robust features*. In ECCV.

Hipiny, I. (2013). *Egocentric activity recognition using gaze*. PhD thesis, University of Bristol.

Joseph., S., Ujir, H. and Hipiny, I. (2017). *Unsupervised classification of Intrusive igneous rock thin section images using edge detection and colour analysis*. In ICSIPA 2017.

Hipiny, I. and Ujir, H. (2015). *Measuring task performance using gaze region*. In CITA 2015.

Ujir, H., Spann, M., and Hipiny, I. (2014). *3D Facial Expression Classification using 3D Facial Surface Normals*. In ROVISP, pg. 245-253.

Ujir, H., and Spann, M. (2014). *Surface Normals with Modular Approach and Weighted Voting Scheme in 3D Facial Expression Classification*. International Journal of Computer & Information Technology, 3(5).